# Designing Socially Assistive Robots

Exploring Israeli and German Designers' Perceptions

Ela Liberman-Pincu *

Ben-Gurion University of the Negev, Beer-Sheva, Israel, elapin@post.bgu.ac.il

Oliver Korn

Affective & Cognitive Institute, Offenburg University, Offenburg, Germany, oliver.korn@acm.org

Jonas Grund

Jonas Grund - Illustration & Comic, Germany, mail@jonasgrund.de

Elmer D. van Grondelle

Delft University of Technology, Delft, The Netherlands, E.D.vanGrondelle@tudelft.nl

Tal Oron-Gilad

Ben-Gurion University of the Negev, Beer-Sheva, Israel, orontal@bgu.ac.il

Socially assistive robots (SARs) are becoming more prevalent in everyday life, emphasizing the need to make them socially acceptable and aligned with users' expectations. Robots' appearance impacts users' behaviors and attitudes towards them. Therefore, product designers choose visual qualities to give the robot a character and to imply its functionality and personality. In this work, we sought to investigate the effect of cultural differences on Israeli and German designers' perceptions and preferences regarding the suitable visual qualities of SARs in four different contexts: a service robot for an assisted living/retirement residence facility, a medical assistant robot for a hospital environment, a COVID-19 officer robot, and a personal assistant robot for domestic use. Our results indicate that Israeli and German designers share similar perceptions of visual qualities and most of the robotics roles. However, we found differences in the perception of the COVID-19 officer robot's role and, by that, its most suitable visual design. This work indicates that context and culture play a role in users' perceptions and expectations; therefore, they should be taken into account when designing new SARs for diverse contexts.

CCS CONCEPTS • Human-centered computing~Interaction design~Systems and tools for interaction design•Human-centered computing~Human computer interaction (HCI)~Empirical studies in HCI•Human-centered computing~Human computer interaction (HCI)~HCI design and evaluation methods

---

* Place the footnote text for the author (if applicable) here.

# 1 INTRODUCTION

## 1.1 Design of Socially Assistive Robots (SARs)

SARs are becoming more prevalent in everyday life [1-3], fulfilling different roles and tasks and establishing varied relationships [4]. The context of the use of SARs can be parsed into four contextual layers: the domain in which the SAR exists, the physical environment, its intended users, and the robot's role; these layers define the human-robot relationship [5]. Even though researchers are working to ensure that SARs follow social norms and expectations [6-8], knowledge is still limited regarding the visual design research of SARs [9,10]. Indeed, the design space for SARs is vast, and narrowing it down to specific use-cases is an extremely demanding task, as a recent interdisciplinary overview of advances in social robotics clearly illustrates [11]. Instead of designing new systems bottom-up and based on user requirements, much research in the field focuses on evaluating users' perceptions of existing off-the-shelf SARs [12-15].

SARs' morphology varies between anthropomorphic (human-like and androids), zoomorphic (animal-like), and technical (machine-like) [16-19]; different robots' tasks and use cases deserve using different morphology [7,20,21]. For example, using a human-like appearance is more suitable in cases where the robot should be perceived as sociable [20]. Still, there is more to design than morphology: designers use different VQs such as shapes, colors, outlines, textures, and dimensions as tools to lead the user to desired behavioral and cognitive responses [22-25], even minor design manipulations were found to affect users' perception and behaviors [26]. This work focuses on basic VQs as understanding the effect of those may also implicate the design of anthropomorphic and zoomorphic robots.

Matching the physical embodiment of the SAR to its task can help reflect the robots' functions and capabilities and improve users' acceptance [16,21,27,28]. In addition, studies conducted among potential users found that the robots' role and context of use affect users' expectations and selections of their robotic appearance [5,29]. Hence, it's evident that the context of use must be incorporated into the design process. Still, manufacturers often utilize the same embodiment for different contexts [5].

In [30], a deconstruction of existing commercial robots' visual and physical components was used to create a design taxonomy. Following this, the authors reconstructed and created 30 new self-designed robots that differ by their three basic VQs: body structure, outline, and color to study the impact and value of the visual design of SARs. Building upon the reconstruction of these robots, it was found that the SARs' VQs impact people's perception of the robot's character as being friendly, childish, innovative, threatening, etc. For example, an A-shape or hourglass structure would be a better choice for the design of a friendly SAR than a V-shape [30]. While users were able to ascribe characteristics to different SARs' VQs, going the other way around and ascribing VQs to express desired characteristics was more of a challenge; participants stated different expectations regarding the robot's characteristics suitable for each context. However, when asked to select VQs that best express these expectations, participants' personal preferences and demographic data were more significant factors affecting their selections [5].

## 1.2 Cultural differences in SARs' perception

Following Hofstede [31] developed the cultural dimensions theory to explain national cultural differences in perceptions and behaviors. This five dimensions model considers power distance, collectivism and individualism, gender roles, uncertainty avoidance, and long and short-term Orientation.

Users' cultural values, beliefs, and habits affect their expectations of robots [32,33]. Previous studies found cultural differences regarding the likability and familiarity of robots [34] as well as the levels of trust, compliance, and comfort in interacting with robots [35-39]. These may affect the design of interaction patterns and cognitive models [19,40-42]. In addition, [43] found cultural differences in robots' role perception; while Japanese participants considered humanoid robots as an extension of humans, UK participants preferred humanoid robots would not perform tasks that require "human-like" qualities, such as empathy, caring, or independent decision making.

Research related to cultural design preference in terms of visual appearance focused mainly on robot morphology [44-46], although researchers have no clear agreement regarding the appropriate design for each culture. For example, [44,46,47] found that participants of Asian cultures (Korean and Japanese) preferred more human-like robots, while Western cultures (US and European) preferred machine-like robots. On the other hand, [45] found an opposite trend comparing US American and Japanese preferences and perceptions. Other studies evaluated the cultural effect on specific visual aspects of robots, such as the choice of materials [47] and dimensions [48]. Although understanding cultural aesthetic preferences is important when designing products for global distribution [49,50], we found no studies on robotic aesthetics perception and culture.

### 1.3 The role of designers

In the design domain, visual features are used to communicate and represent concepts and ideas [51]; their selection can directly affect the user's actions [52]. Designers rely on their experience, habits, intuitive feeling, and inspiration from artistic works to come up with the visual qualities for a new product [53]. Therefore it is reasonable that cultural background [50], as well as the level of experience [51], affect their design thinking, processes, and final outcomes. Although the current ideal is a "user-centered" or "human-centered" process, also referred to as "co-design," where users are actively involved in the design process, this ideal is rarely achieved. Financial and temporal constraints often result in designers' drafts and prototypes making their way into the final solution. A commendable exception is the design of a range of sketches for virtual assistants for users with impairments [52] performed by two of the co-authors. In this study, we explore the perception and preferences of design students and professional designers of two cultures, Israeli and German.

## 2 STUDY DESIGN

### 2.1 Aim and Scope

Previous studies [5,30] evaluated users' perceptions and preferences. In this study, we aim to understand the effect of cultural background on professional designers' perceptions; perceptions of a robot's desired characteristics in a context of use, and perceptions of the most suitable VQs to express it. The following questions were addressed: (1) What affects designers' selections when designing SARs in different contexts? (2) How does the cultural background impact the perception of SARs' roles and desired characteristics? (3) Are there shared perceptions regarding the meaning of visual qualities among designers of different cultures? Three research hypotheses were formulated and are detailed below.

**Hypothesis 1 [H1]** Professional designers consider the context of use (and branding) in their design; it is part of their professionalism. Therefore, we expect to find common links in their approach to the context of use, the desired characteristic, and the selected VQs.

**Hypothesis 2 [H2]** Designers from different cultures perceive robotic roles differently. Studies have shown that cultural differences affect users' perceptions of robots' roles and functions [19,32-43]; we assume that professional designers are influenced by their cultural background.

**Hypothesis 3 [H3]** Designers from different cultures have different perceptions regarding the meaning of visual qualities attributes. As [49] stated: "Understanding the culture of others may be an essential element in creating products that people in other parts of the world appreciate." We assume the cultural background of the designers participating in our study will affect their design language.

## 2.2 Four use cases

The context of use of SARs can be deconstructed into four contextual layers: the domain in which the SAR exists (e.g., healthcare, educational, entertainment, etc.), the physical environment (i.e., indoor or outdoor, personal or public), its intended users (i.e., professional or nonprofessional, demographics, needs and abilities), and the robot's role (by abstract roles and by a human-robot hierarchy) [5]. Therefore, we defined four SAR use cases that differ by their contextual layers:

**A service robot for an Assisted Living/retirement residence facility (ALR)** aims to roam the lobby and be used by the facility residents to register for various classes and activities. In addition, it provides information and helps communicate (via video calls and chats) with staff members.

**A Medical Assistant Robot (MAR) for a hospital environment** aims to assist the medical team, especially when social distancing is required. Through it, the medical team can communicate in video calls with isolated patients and bring equipment, food, and medicine into patients' rooms.

**A COVID-19 Officer Robot (COR)** aims to ensure passersby comply with Covid-19 restrictions like social distancing or wearing a face mask.

**A Personal Assistant Robot (PAR)** for home/domestic use seeks to assist users with daily tasks, recommend activities at home and outside, and remind them of their duties and appointments. In addition, the robot allows users to watch videos, listen to music, play, and have video chats with family and friends.

## 2.3 Study design

*2.3.1 Online Questionnaire Design*

Using Qualtrics, we designed an online questionnaire where participants (professional designers and design students from Israel and Germany) were randomly presented with one of the four use cases. First, they were asked to select relevant words out of a word bank to define the robot's desired characteristics. The word bank contained twelve words based on previous studies related to SARs' perception [55-57] that were found relevant to the use cases: innovative, inviting, cute, elegant, massive, friendly, authoritative, aggressive, reliable, professional, intelligent, and threatening. In addition, respondents had the option of adding their own adjectives. Following, they were asked to design the robot by selecting three types of VQs based on a VQs' taxonomy [30] that, in their opinion, best expresses the desired characteristics that they have chosen: body structure (A-shape, Diamond, Hourglass, Rectangle, or V-shape), outline (Chamfered or Rounded), and color combination (Dark colors, White and blue combination, or White). The questionnaire was translated into Hebrew and German using a two-way translation procedure. Figure 1 illustrates the questionnaire design.

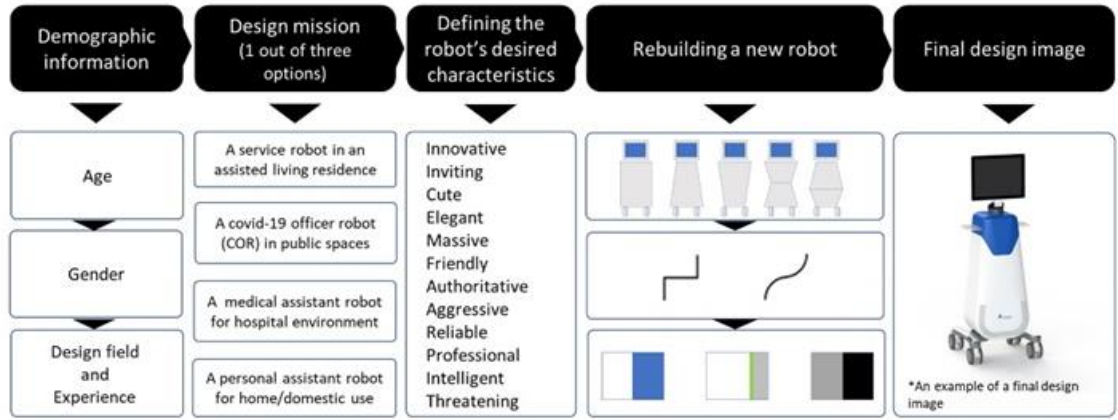

Figure 1: questionnaire design.

*2.3.2 Questionnaire distribution*

We contacted design faculties of universities, industrial design firms, and professional associations of designers in Israel and Germany. We asked them to distribute the questionnaire to their members, coworkers, teaching staff, and students via their mailing lists and social media platforms like Facebook and LinkedIn.

## 3   RESULTS

In total, we collected data from 200 professional designers and design students aged 18-72 (M=31.3, SD=10.4) from Israel and Germany. Table 1 Summarizes the respondents' gender and work experience by use case. Respondents' professional backgrounds as designers varied between 3D designers (including product design, fashion, and art; 52%), and 2D designers (including graphic design, illustration, and game design; 48%), and two levels of professional experience: students (52%), and professional designers (48%) as detailed in Table 2. Professional experience and age were highly correlated (0.7); therefore, in our following analyses, where relevant, we used the factor of professional experience rather than age.

Table 1. Respondents' data by use case

| Use case | Israelis | | | Germans | | | |
|---|---|---|---|---|---|---|---|
| | Gender Males | Females | Age M(SD) | Gender Males | Females | Not disclosed | Age M(SD) |
| ALR | 8 | 20 | 32 (10.3) | 8 | 11 | 2 | 28.2 (10.6) |
| MAR | 9 | 24 | 32.2 (8.9) | 9 | 14 | 0 | 28.7 (10.4) |
| COR | 16 | 15 | 35.5 (12.5) | 8 | 14 | 1 | 30.8 (11.3) |
| PAR | 8 | 12 | 31.7 (7.6) | 11 | 10 | 0 | 28.4 (9.3) |
| **Total** | **41** | **71** | **33 (10.1)** | **36** | **49** | **3** | **29.0 (10.3)** |

Table 2: Respondents' professional backgrounds

| Use case | Design field | | Professional experience | |
|---|---|---|---|---|
| | 3D designers | 2D designers | Design students | Professional designers |
| ALR | 24 | 25 | 31 | 18 |
| MAR | 31 | 25 | 27 | 29 |
| COR | 32 | 22 | 23 | 31 |
| PAR | 17 | 24 | 23 | 18 |
| **Total** | **104** | **96** | **104** | **96** |

We analyzed the results to find the factors affecting the participants' selections of desired characteristics for the SARs (section 3.1) and the selection of visual qualities (section 3.2). Following, we analyzed the connections between the selections to see if the designers share the same language to express different characteristics (section 3.3). In addition, 91 respondents (59 Israelis and 32 Germans) contributed additional comments (one or more); we excluded greetings and vague comments. The remaining 121 comments were analyzed using thematic analysis (section 3.4). The following paragraphs present the designers' selections and relevant statements as they provide insights to understand the results better.

### 3.1 Selection of desired characteristics for the SARs

We evaluated the effect of the context and participants' characteristics, cultural background (Israeli/German), design field (3D or 2D), professional experience (professional/student), and gender on their selections of desired characteristics across all use cases. Regardless of the use context and the participants' origin, the most selected word was *Reliable*; 78.5% of the designers marked this word as a required characteristic. It appeared as one of the top selected words in all four use contexts.

The context of use was found to affect the designers' selection of describing words. In addition, we found that the participants' cultural background, professional experience, and gender affected their expectations and perception of the robots' desired characteristics. All in all, we found statistically significant effects for seven words: innovative, inviting, elegant, friendly, authoritative, professional, and intelligent. Table 3 summarizes the results; We excluded the word *Aggressive*, which was selected only in 1% of the cases, and the words *Massive* and *Threatening*, which were selected only in 2%. A table containing all words and selections by the different factors can be found in Appendix A.

Table 3: Factors affecting the designers' selection of words

| | Innovative | Inviting | Elegant | Friendly | Authoritative | Professional | Intelligent |
|---|---|---|---|---|---|---|---|
| Context | $p < .01$ | $p < .01$ | $p < .01$ | $p < .01$ | $p < .01$ | | |
| Culture | | $p < .05$ | | $p < .05$ | $p < .01$ | | $p < .01$ |
| Gender | $p < .05$ | | | | | $p < .05$ | |
| Design Field | | $p < .01$ | | $p < .01$ | $p < .05$ | $p < .05$ | $p < .01$ |
| Experience | | | | | | $p < .01$ | |

Light gray boxes represent a significant level of $p < .05$, and dark gray boxes represent a significant level of $p < .01$.

*3.1.1 Desired characteristics by context*

The use context presented to the designers was found to significantly affect ($p < .01$) their selection of five describing words: *Innovative, Inviting, Elegant, Friendly*, and *Authoritative*. The word *Innovative* was mainly ascribed to the MAR use case; 53.6% of the designers thought a medical robot should look innovative, compared to 16.3% in the ALR use case. The word *Inviting* was selected mainly for the ALR use case; 76% of the designers selected it for this context compared to only 39% for the context of COR. The designers found the word *Elegant* relevant mainly to the context of PAR (43% compares to 23% in the overall data). The context of COR was perceived as the most *Authoritative* (48% compared to 19.5% in the overall data) and the least Friendly (61% compares to 77.5% in the overall data).

Still, the top selected words did not differ greatly between the four use cases. The word *Reliable* was selected as a relevant characteristic for all four use cases, the word *Friendly* was selected for all except for the COR use case, the word *Professional* was selected for MAR and COR, and *Inviting* was chosen as a required characteristic for ALR. Table 4 presents the most selected words for each context and their rate. The complete table is in Appendix A.

Results were mostly similar among the Israeli and German designers, except for the COR use case; a detailed analysis of the different use cases by cultural background is presented in section 3.1.3.

Table 4. Most selected words (over 65%) for each context and their rate.

| Case study | Most selected words |
|---|---|
| ALR | Reliable (88%) |
| | Friendly (88%) |
| | Inviting (76%) |
| MAR | Friendly (86%) |
| | Reliable (84%) |
| | Professional (70%) |
| COR | Reliable (70%) |
| | Professional (68%) |
| PAR | Friendly (76%) |
| | Reliable (71%) |

*3.1.2 Cultural background differences*

Excluding the context, the respondent's culture significantly affected their selection of four describing words; German designers were more likely than Israelis to select three words: *Inviting, Friendly*, and *Intelligent,* while Israeli designers showed a higher tendency to select the word *Authoritative*. Table 5 summarizes these results; Figure 2 illustrates the selection of words by culture in a radar chart.

Table 5. Culture-related four describing words.

| | Israelis | Germans | |
|---|---|---|---|
| Inviting | 45% (*N*=50) | **63% (*N*=55)** | $X^2 (1, N = 200) = 6.3, p < .05$ |
| Friendly | 71% (*N*=80) | **85% (*N*=75)** | $X^2 (1, N = 200) = 5.38, p < .05$ |
| Authoritative | **30% (*N*=34)** | 6%  (*N*=5) | $X^2$ (1, *N* = 200) = 19.15, *p* < .01 |
| Intelligent | 38% (*N*=42) | **63% (*N*=55)** | $X^2$ (1, *N* = 200) = 12.33, *p* < .01 |

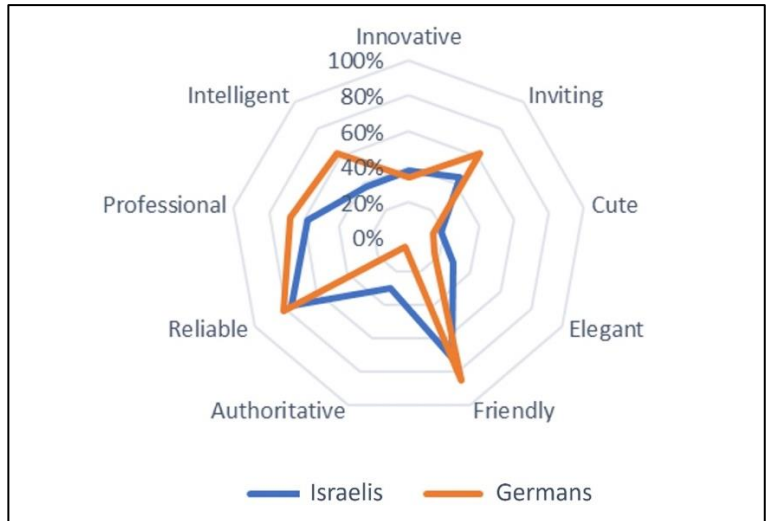

Figure 2: A comparison of Israeli and German designers' selection of describing words presented in a radar chart.

*3.1.3 Cultural differences by use context*

Here we look at each one of the four use cases separately.

### *A Medical Assistant Robot (MAR) for a hospital environment*

The top selected words for the context of MAR were *Friendly*, *Reliable*, and *Professional*. German respondents also tend to select the word *Intelligent* (65%), while only 30% of the Israeli designers chose it. In addition, both Israeli and German designers suggested additional words for this context. Among the Israeli designers, we identified two groups of words related to two different medical roles: **Nursing roles** (*Calming, Comforting*, and *Compassionate*) and **Expert roles** (*Practical, Clean*, and *Purposeful*). German designers contributed three additional words that are related to **Comforting roles** (*Gentle, Familiar*, and *Funny*). Table 6 presents the Most selected words by Israeli and German designers for the context of MAR and their rate.

Table 6. Most selected words by Israeli and German designers (over 65%) for the context of MAR and their rate.

| Most selected words | Most selected words by Israeli designers | Most selected words by German designers |
|---|---|---|
| Friendly (86%) | Friendly (81%) | Friendly (96%) |
| Reliable (84%) | Reliable (81%) | Reliable (87%) |
| Professional (70%) | Professional (72%) | Professional (65%) |
| | | Intelligent (65%) |

We found statistically significant relations among the Israeli respondents for the context of MAR and two describing words: *Innovative* ($X^2$ (1, $N$ = 112) = 8.34, $p < .01$) and *Professional* ($X^2$ (1, $N$ = 112) = 5.27, $p < .05$); 59% of the Israelis designers selected the word *Innovative* for the design of a medical robot compared to 38% in the overall data, and 72% of them selected the word *Professional* compared to 58% in the overall data.. Figure 3 illustrates the two cultural groups' chosen words for the context of MAR in a radar chart.

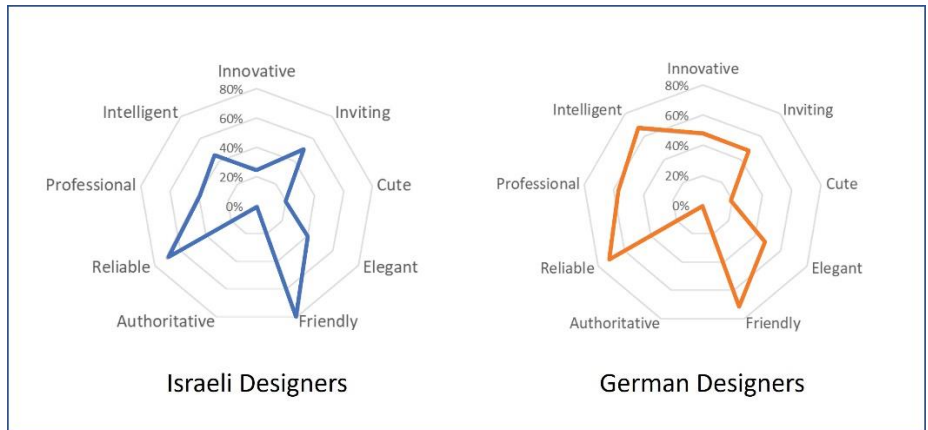

Figure 3: Respondents' assigned characteristics for the context of MAR by culture

*A service robot for an Assisted Living/retirement residence facility (ALR)*

Table 7. presents the most selected words by Israeli and German designers (over 65%) for the context of ALR and their rate. The top selected words for the context of A service robot for an Assisted Living/retirement residence facility (ALR) were: *Reliable, Friendly*, and *Inviting*. German designers also indicated *Professional* as a desired characteristic 81% of them selected it, while only 42% of the Israeli designers did. In addition, four Israeli designers suggested additional words for this context: *Familiar, Protecting*, *Gentle*, and *Easy to use*; one designer explained in detail:

> "In my opinion, older people may feel intimidated by technology; their fears must be considered. We should think about how to make them feel more comfortable by designing robots more similar to the things that are familiar to them."

None of the German designers contributed additional words for this context, but one did add a detailed remark regarding the design process for older adults:

> "Seniors have a different approach to technology; computers often overwhelm them. Young generations are digital natives, are automatically attracted to something like robots, or are more likely to deal with and operate them. This is different with older people. I would definitely go to retirement homes and talk to the seniors. They know best how this robot has to be designed."

Table 7. Most selected words by Israeli and German designers (over 65%) for the context of ALR and their rate.

| Most selected words | Most selected words by Israeli designers | Most selected words by German designers |
| --- | --- | --- |
| Reliable (88%) | Reliable (83%) | Reliable (95%) |
| Friendly (88%) | Friendly (83%) | Friendly (90%) |
| Inviting (76%) | Inviting (66%) | Inviting (86%) |
| | | Professional (81%) |

We found statistically significant relations between the context of ALR and two describing words: *innovative* ($X^2$ (1, $N$ =200) = 10.9, $p < .01$) and *inviting* ($X^2$ (1, $N$ = 200) = 13.78, $p < .01$). 76% of the designers (Israelis and German) selected the word *Inviting* for the design of an ALR compared to 45% in the overall data. In addition, designers were less likely to ascribe the word *innovative* to the design of an ALR; only 16% of them did, compared to 42% in the overall data. Figure 4 illustrates the two cultural groups' chosen words for the context of ALR in a radar chart.

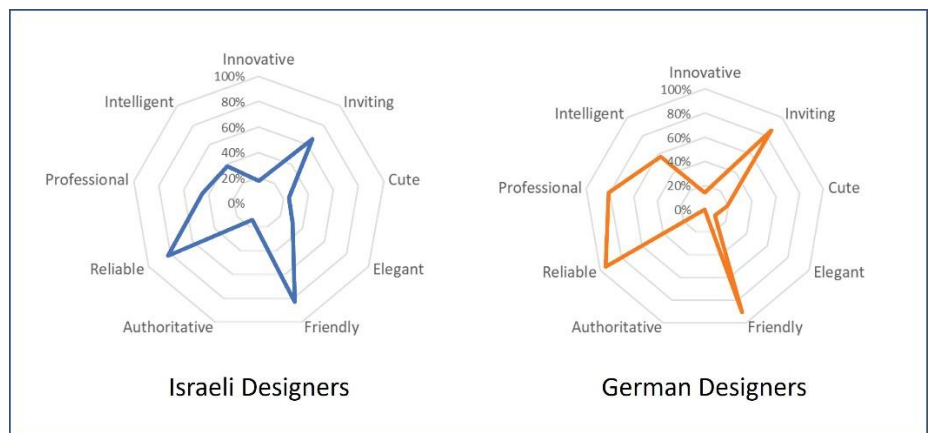

Figure 4: Respondents' assigned characteristics for the context of ALR by culture.

### *A Personal Assistant Robot (PAR) for home/domestic use*

The top selected words for the context of A Personal Assistant Robot (PAR) were identical for both cultures: *Friendly* and *Reliable*. German designers were also more likely to select the word *Intelligent*; 67% of them did, compared to only 50% of the Israeli designers. In addition, six designers suggested additional words for this context; Israeli designers contributed: *Soft* and *Minimalistic*, and German designers contributed: *Reserved* and *Discreet*. in addition, some asked for it to be "not too big" and "Likeable, with a certain character." Table 8. Presents the most selected words by Israeli and German designers (over 65%) for the context of PAR and their rate.

Table 8. Most selected words by Israeli and German designers (over 65%) for the context of PAR and their rate.

| Most selected words | Most selected words by Israeli designers | Most selected words by German designers |
|---|---|---|
| Friendly (76%) | Friendly (80%) | Friendly (71%) |
| Reliable (71%) | Reliable (70%) | Reliable (71%) |
| | | Intelligent (67%) |

We found statistically significant relations between the context of PAR and the selection of the describing words: *Elegant* ($X^2$ (1, $N$ =200) = 11.9, $p$ <.01) and *Professional* ($X^2$ (1, $N$ = 200) = 4.1417, $p$ <.05). The designers (Israeli and Germans) were less likely to select the word *Professional* for this context (49% compared to 66% in the overall data) and more likely to choose the word *Elegant* (44% compared to 18% in the overall data). In addition, no designer selected the word *Authoritative* for this context (compared to 32.5% in the overall data). Figure 5 illustrates the two cultural groups' chosen words for the context of PAR in a radar chart.

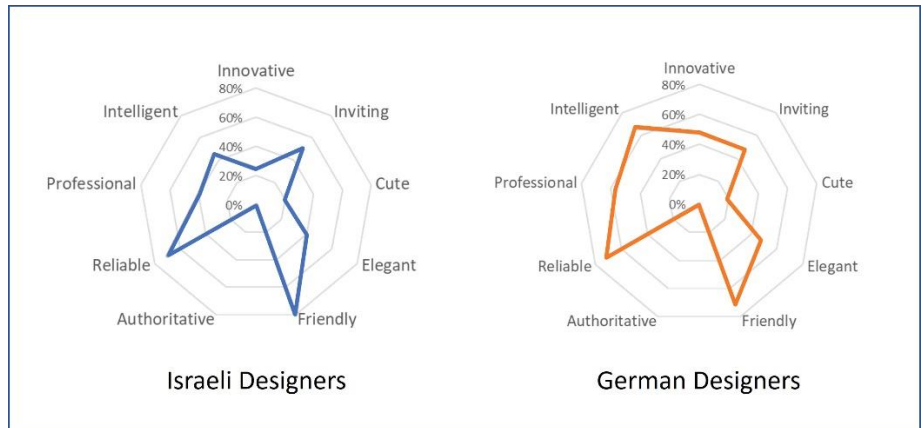

Figure 5: Respondents' assigned characteristics for the context of PAR by culture.

### A COVID-19 Officer Robot (COR)

The top selected words for the context of Covid-19 Officer Robot (COR) in the overall data were Reliable (70%) and Professional (69%). However, looking at each culture separately reveals the different perceptions of this context's role. 68% of the Israeli designers selected the word Authoritative as one of the most suitable characteristics, while only 22% of the German designers did. On the other hand, the German designers aimed for a friendlier robot; Friendly was the most selected word for the context of COR by the German designers (83% compared to 45% among the Israeli designers). One German designer suggested the additional word Sovereign. None of the Israeli designers suggested additional words for this context. Table 9. Presents the most selected words by Israeli and German designers (over 65%) for the context of COR and their rate.

Table 9. Most selected words by Israeli and German designers (over 65%) for the context of COR and their rate.

| Most selected words | Most selected words by Israeli designers | Most selected words by German designers |
|---|---|---|
| Reliable (70%) | Authoritative (68%) | Friendly (83%) |
| Professional (69%) | Reliable (68%) | Reliable (74%) |
|  | Professional (68%) | Professional (70%) |

Since the results showed these two cultures do not share the same perception of this robot's role and character, we conducted the chi-square test of independence separately. Among the Israeli designers, we found statistically significant relations between the context of COR and four describing words: *Authoritative, Inviting, Friendly, and Elegant;* The designers were aiming for a COR to look more *Authoritative ($X^2$ (1, N = 112) = 28.34, p < .01)* and less *Inviting ($X^2$ (1, N = 112) = 9.11, p < .01), Friendly ($X^2$ (1, N = 112) = 14.49, p < .01), and Elegant ($X^2$ (1, N = 112) = 6.94, p < .01)* than the three other contexts. Still, some Israeli designers thought differently; one of them explained his selection of the words: *Friendly*, *Inviting*, *Reliable*, and *Professional*:

> "In my opinion, we should design a robot, not human-like, but as close as possible to a human figure, so it will not create a certain reluctance or fear when facing an adult or a 5-year-old child… …It should be as friendly as possible. Not threatening or aggressive, it should give even the most suspicious person a sense of confidence and friendliness."

Among the German designers, we found statistically significant relations between the context of COR and the word *Authoritative* ($X^2$ (1, N = 88) = 13.12, p < .01). The German designers used this word only for this context. Figure 6 illustrates the two cultural groups' chosen words for the context of COR in a radar chart.

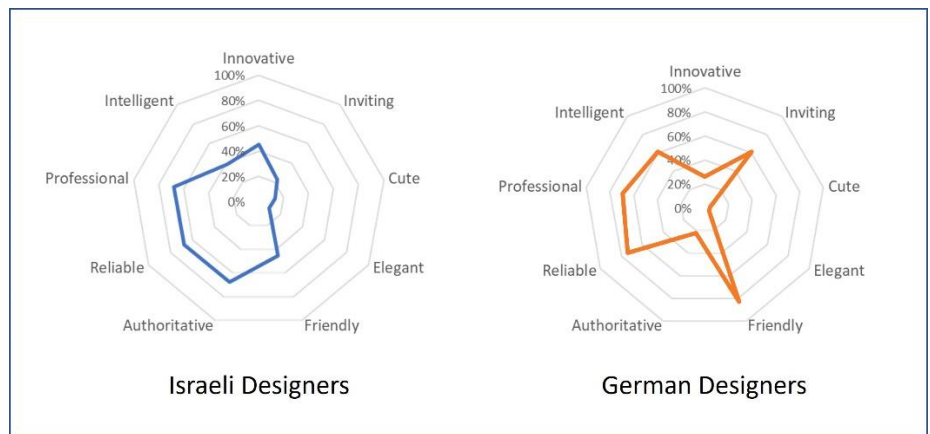

Figure 6: Respondents' assigned characteristics for the context of COR by culture.

*3.1.4 Gender differences*

We analyzed the effect of the respondents' demographic information on their word selections using a Chi-square of independence. Since age was highly correlated with professional experience, we report the impact of the second in the next section. Female designers expressed their desire for innovative and professional robots. 42.5% selected the word *Innovative* compared to only 27% of the male designers. The relation between these variables was significant ($X^2$ (1, N = 197) = 4.69, p < .05). 69% of the female designers selected the word *Professional* compared to only 53% of the male designers ($X^2$ (1, N = 197) = 5.09, p < .05).

*3.1.5 Differences related to the professional design field and experience*

We analyzed the effect of the respondents' design field and professional experience on their word selections using a Chi-square test of independence. Results showed that respondents' design field, and professional experience affected their selections of six describing words: *Innovative, Inviting, Friendly, Authoritative, Professional*, and *Intelligent*.

Respondents' design field was found to affect their perceptions. 3D designers (product, fashion, and art) were more likely to select an authoritative character for their robot than 2D designers (graphic, illustration, and game design), 26% compared to 12.5% ($X^2$ (1, $N$ = 200) = 5.76, $p$ < .05). In addition, they were less likely to select *Inviting*, 42% compared to 64% ($X^2$ (1, $N$ = 200) = 9.03, $p$ < .01), *Friendly*, 69% compared to 86% ($X^2$ (1, $N$ = 200) = 8.5, $p$ < .01), *Professional*, 56% compared to 70% ($X^2$ (1, $N$ = 200) = 4.19, $p$ < .05) and *Intelligent*, 38% compared to 59% ($X^2$ (1, $N$ = 200) = 8.74, $p$ < .01).

Design students were more likely to desire a professional-looking robot (71%) than professional designers (53%). The relation between these variables was significant ($X^2$ (1, $N$ = 200) = 6.78, $p$ < .01).

### 3.2 Selection of visual qualities

After selecting the characteristics of the SAR, designers were asked to choose the most suitable visual qualities for the context of use they had. Some VQs were selected more frequently than others, regardless of the context, culture, or other factors. For example, most designers (87.5%, Israelis: 86%, Germans: 90%) preferred rounded edges over chamfered ones. Only 11% of them (Israelis: 8%, Germans 14%) chose the dark color scheme, and most (55.5%, Israelis 56%, Germans 54.5%) preferred the white color scheme. The two most selected structures were the A shape (31%) and the Diamond shape (30.5%). We found significant effects of the context on the respondent's selections of structure ($X^2$ (12, $N$ = 200) = 24.98, $p$ <.05) and color ($X^2$ (6, $N$ = 200) = 24.64, $p$ <.01). Cultural design differences were only found in the case study of COR.

The following paragraphs present the VQs selections of all respondents together by context; the context of COR is discussed in detail in section 3.2.4. The respondents' gender and professional experience did not affect any of the VQs' selections. A table containing all VQs' selections by the different factors can be found in Appendix B.

*3.2.1 Body structure*

Out of the five options for the body structure (A shape, Diamond, Hourglass, Rectangle, and V-shape), three were selected as the final design (most selected) for the four contexts. For the design of PAR, most designers (51%) selected the diamond shape (compared to 30.5% in the overall data). Even though the Diamond was the preferred structure for this context among both cultures, Israeli designers showed an even higher tendency toward it; 65% selected it compared to 38% of the German designers. For the design of MAR, most of the designers (39%) selected the A-shape (compared to 31% in the overall data). The two cultures' selections were highly similar. For the design of ALR, most designers (37%) selected the A-shape (compared to 31% in the overall data). Again, the two cultures' selections were highly similar. For the design of COR, most designers (35%) selected the V-shape structure (compared to 20% in the overall data). However, as detailed in the following section, we have found significant differences between the Israeli and German selections. Figure 7 illustrates the designers' structure selections by context.

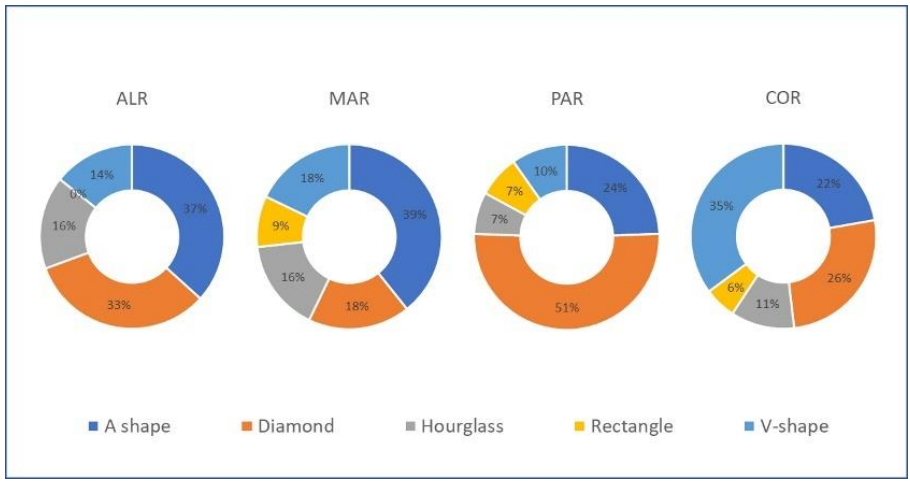

Figure 7: designers' structure selections by context

*3.2.2 Outline*

Most designers selected a rounded outline for all four contexts. No significant relations were found between the context and the outline selection. Figure 8 illustrates the designers' outline selections by context.

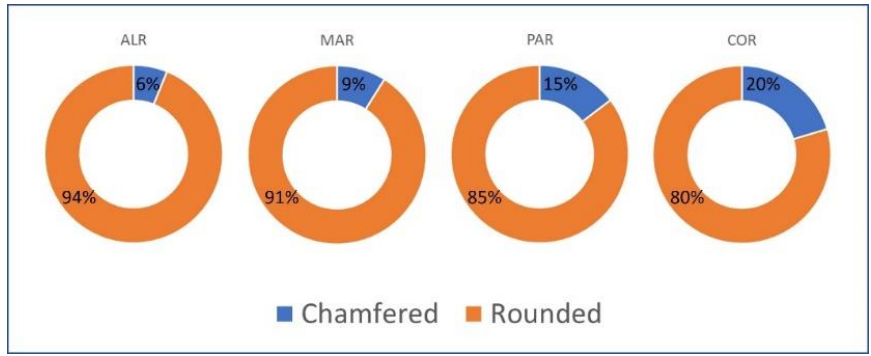

Figure 8: the designers' outline selections by context.

*3.2.3 Color*

The context of use affected the designers' tendency to select a color combination for their robot, the relationship between these two variables was significant ($X^2$ (6, $N$ = 200) = 24.64, $p$ <.01). White was the most selected color out of the three options; 55.5% of the respondents chose it. Furthermore, it was the preferred color for the design of COR (67%), ALR (63%), and MAR (55%). For the use context of PAR, there was no explicit agreement among the designers; the most selected color was the white and blue combination (39% compared to 33.5% in the overall data), followed by the white color (32% compared to 55.5% in the general data) and the dark combination (29% compared to 11% in the general data). Figure 9 illustrates the designers' color selections by context. The comments also supported these results; three of the designers in this context (two German and one Israeli) mentioned they would like to have more color options:

"As a customer, I would probably like to be able to choose the color selection myself because when buying a new item in the household, you always pay attention to the environment (many of our furniture are black; hence I would tend towards black objects for new purchases)."

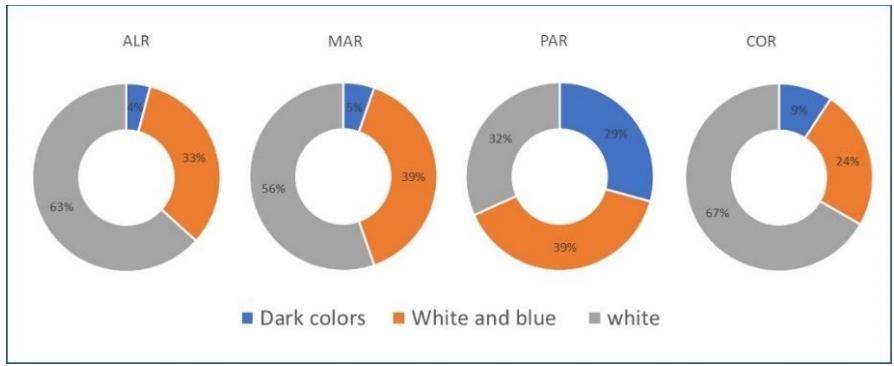

Figure 9: The designers' color selections by context.

### 3.2.4 Cultural differences in the context of COR

Israeli and German designers' selections shared resemblance for the PAR, MAR, and ALR designs. However, we found cultural differences in the design of COR; the Israeli designers preferred the V-shape structure for this context (42%), while the German designers selected a Diamond structure (38%). Figure 10 compares Israeli and German designers' body structure selection in the COR context. One of the Israeli designers explained her selection of the V-shape structure:

"Inspectors are antagonists because their job is to enforce the law. Therefore, they must be designed in a masculine form to avoid encouraging vandalism in places where society is patriarchal."

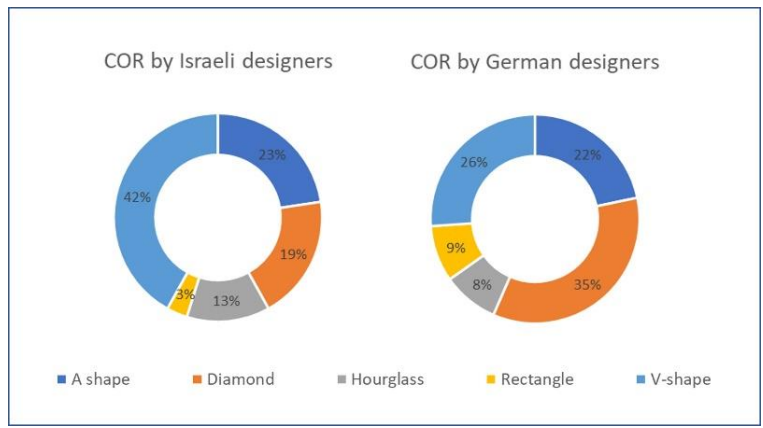

Figure 10: Israeli and German designers' body structure selection in the COR context.

Although rounded edges were the preferred outline across all four contexts and the two cultures, a Chi-square test of independence found a significant relationship between the presented use context and the selection of edge type ($X^2$ (3, $N$ = 112) = 8.67, $p$ <.05) among Israeli designers. The Israeli designers showed a higher tendency to select chamfered edges in the case of COR (29% compared to 14% in their overall data). On the other hand, German designers' selections were similar to their overall data (9% compared to 10%). Figure 11 compares Israeli and German designers' outline selection in the COR context.

Both Israeli and German designers mainly selected the white color for the context of COR. However, The German designers' selections were more varied; 57% of them selected white, 30% preferred a combination of white and blue, and 13% chose a dark color scheme. On the other hand, the Israeli designers were more similar; 75% selected white. Figure 12 compares Israeli and German designers' color selection in the COR context.

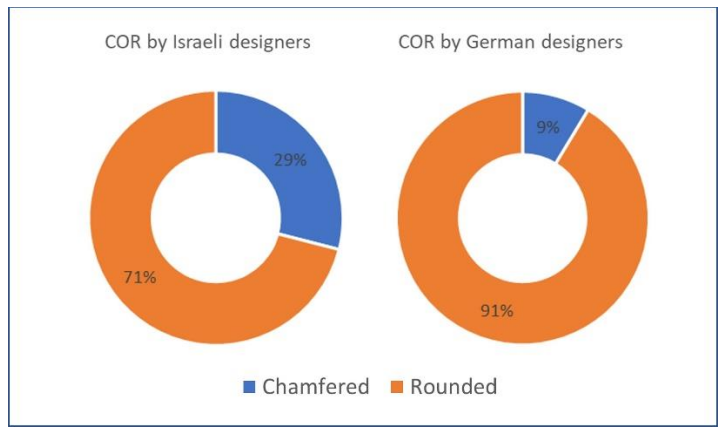

Figure 11: Israeli and German designers' outline selection in the COR context.

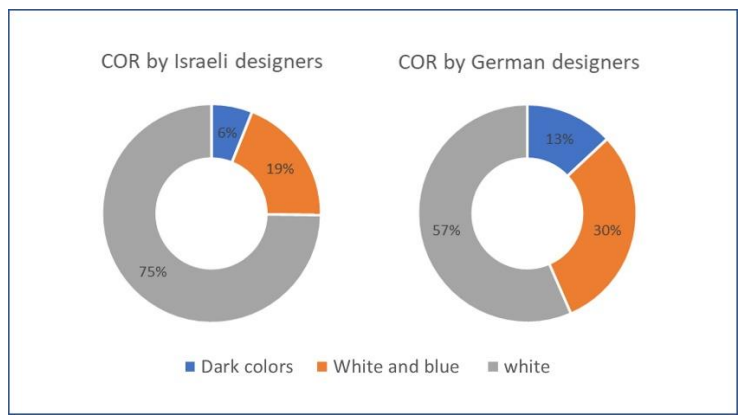

Figure 12: Israeli and German designers' color selection in the COR context.

*3.2.5 The "preferred" design by context and culture*

We summarized the designers' selections to create the "preferred" design for each context by culture. We found that Israeli and German designers' selections were highly correlated when designing MAR, PAR, and ALR. MAR's and ALR's final designs were similar. However, the design of COR differed between the two cultures. Table 10 summarizes the designers' selections of VQs and presents illustrations of the final designs by cultures and contexts.

Table 10: Designers' selection of Visual qualities.

|  |  | COR | MAR | PAR | ALR |
|---|---|---|---|---|---|
| Israelis | Structure<br>Outline<br>Color<br>Illustration | V shape (42%)<br>Rounded (71%)<br>White (74%)<br>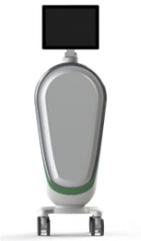 | A shape (38%)<br>Rounded (88%)<br>White (56%)<br>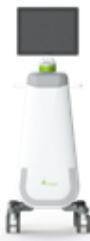 | Diamond (65%)<br>Rounded (90%)<br>**White and blue (40%)**<br>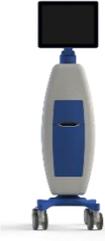 | A shape (38%)<br>Rounded (97%)<br>White (52%)<br>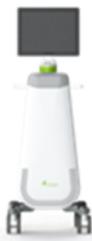 |
| Germans | Structure<br>Outline<br>Color<br>Illustration | Diamond (35%)<br>Rounded (91%)<br>White (57%)<br>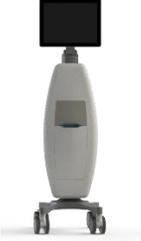 | A shape (39%)<br>Rounded (96%)<br>White (57%)<br>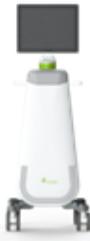 | Diamond (38%)<br>Rounded (81%)<br>**White and blue (38%)**<br>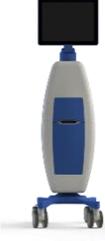 | A shape (38%)<br>Rounded (90%)<br>White (76%)<br>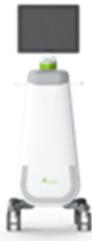 |

### 3.3 Characteristic and VQs selection

We found significant relationships between some of the characteristics the designers chose and the VQs they selected. For example, to express an authoritative robot (regardless of the use context and culture), the most used structure was the V-shape (46%); the relationship between the two variables is significant ($X^2$ (4, $N$ = 200) = 22.26, $p < .01$). In addition, the designers showed significant ($X^2$ (1, $N$ = 200) = 7.65, $p < .01$) higher tendency to select chamfered edges (26% compared to 12.5% in the overall data). To achieve an inviting appearance, the designers avoided using dark colors (6% compared to 11% in the overall data) and showed a little higher tendency to select a white and blue combination. The relation between these variables was significant, $X^2$ (2, $N$ = 200) = 6.36, $p < .05$). Furthermore, designers were significantly ($X^2$ (1, $N$ = 200) = 9.3, $p < .01$) more likely to select rounded edges outline (94% compared to 87.5% in the overall data). Figures 13-15 illustrate the relations between the selected characteristics and VQs' selections.

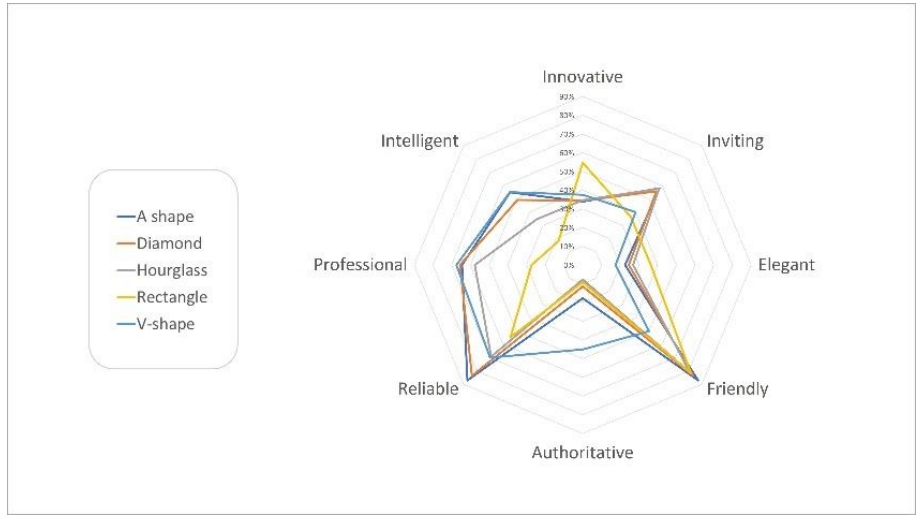

Figure 13: Relations between selected characteristics and structure selection.

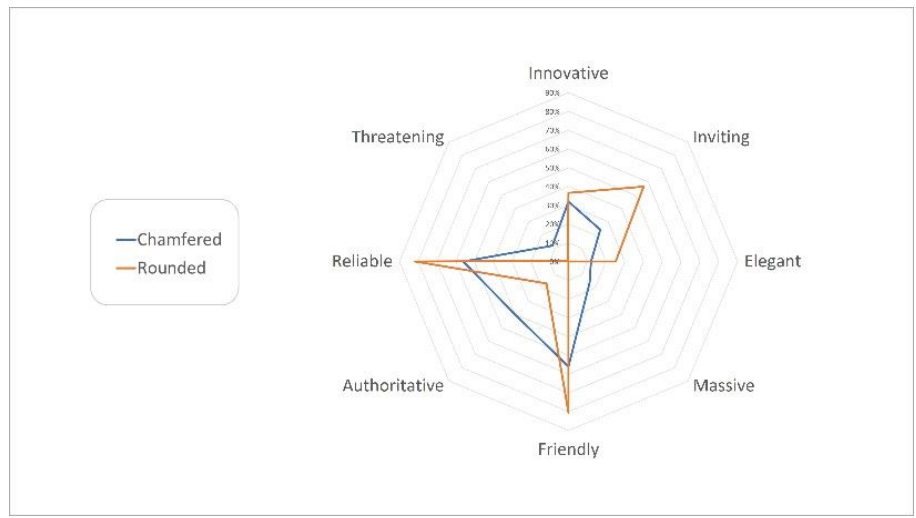

Figure 14: Relations between selected characteristics and outline selection.

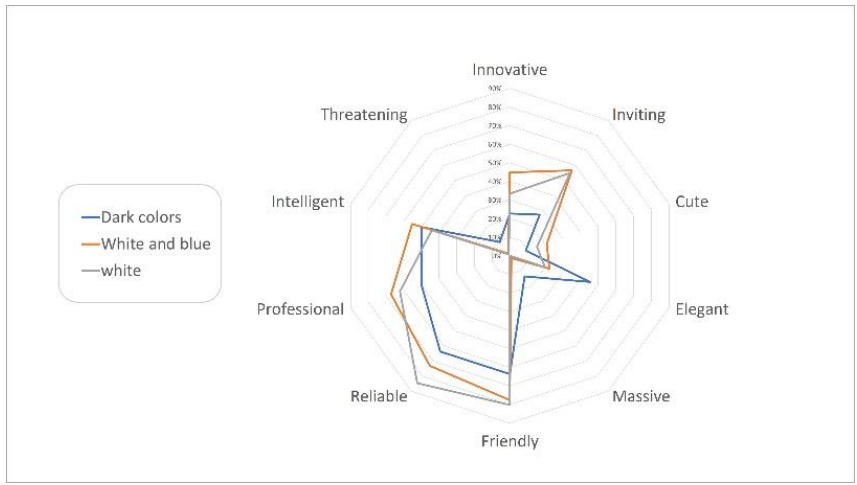

Figure 2: Relations between selected characteristics and color selection.

### 3.4 Thematic analysis

Ninety-one respondents contributed additional comments (one or more). We excluded greetings and vague statements and remained with 121 constructive comments. A thematic analysis of participants' comments revealed five themes (based on [30]: General appearance, The use of main components, Dimensions, Use context, and Supportive interaction. Each is divided into subcategories, as detailed in Table 11.

Table 11: A thematic analysis of participants' comments

| Theme | Sub-themes | Israelis | Germans |
| --- | --- | --- | --- |
| General Appearance | Structure | (n=4) | (n=2) |
|  | Color | (n=9) | (n=10) |
|  | Morphology | (n=4) |  |
| The use of main components | Screen | (n=19) | (n=10) |
|  | Wheels | (n=19) | (n=7) |
| Dimensions | Heights | (n=5) |  |
|  | proportions | (n=4) | (n=2) |
| Use context | Environment conditions | (n=3) | (n=2) |
|  | robot role | (n=3) | (n=2) |
| Supportive interaction | Robots' characteristics | (n=11) | (n=1) |
|  | Interaction and usability | (n=2) | (n=1) |
|  | Graphics and add-ons | (n=1) |  |

The most common theme was using the main components; 55 designers commented regarding the screen and wheels design (regardless of use context). Most of the comments regarding the wheels suggested hiding them, as two explained:

"I would start with hiding the wheels inside the robot's body, creating an enigmatic element in the robot's mobility."

"I would hide the wheels ("floating" effect) to make it look more futuristic and intelligent."

The most frequent comment regarding the screen was a suggestion to make it part of the body, some suggested designing it differently, and one preferred to have a more human-like head instead of a screen.

Fourteen designers asked for more color options (nine Israelis and five Germans), and some of them added color suggestions: green, light gray, orange, and mustard. And four were asking for different structures; two of them suggested more amorphic shapes, as one explained:

"I would use shapes and structures of nature; these are rooted in our mind and memories, expressing empathy and calmness, making us feel in a familiar, safe place even unconsciously."

Eleven designers commented on the robots' dimensions suggesting different heights and body-screen proportions. In addition, three designers asked for more human-like shapes for the cases of COR and ALR. One designer asked for a pet-like figure for the case of PAR.

## 4 CONCLUSION AND FUTURE WORK

In this study, we sought to investigate cultural differences between Israeli and German designers regarding the design perception of SARs in different use contexts. We found that the context affected the designers' desired characteristics of the robot and the selection of its visual qualities. The selected desired characteristics correlated with previous findings of a study conducted among users (not professional designers)[5]. This means that users and designers share the same expectations of SARs in different contexts. But while the designers' selected characteristics guided them in the design process using the three visual qualities to evoke feelings and create a certain character in the robot, the users' selection of VQs seemed to be based more on their personal preferences, and they couldn't link characteristics with VQs.

Our results indicate that Israeli and German designers share similar perceptions of visual qualities' meaning and most of the robotics roles. These findings correlate with [49] arguing that Israel and Germany belong to the same cultural group of "Meritocrats", and share similar aesthetic preferences. However, we found differences in the perception of the COVID-19 Officer Robot's role and, by that, its most appropriate visual design. In order to design an authoritative-looking COR, Israeli designers tend to select the V-shape structure, and the German designers who desired a friendly and reliable-looking COR selected the diamond structure. These correlate with previous findings that linked these VQs with users' perceptions [30].

The use case of PAR has demonstrated once again the importance of allowing customization in the design of personal robots [58]. The designers' color selections for that use case varied across all three options. Furthermore, three designers added comments regarding the need to have more color options to allow matching the design of the robot to the customer's home. Most designers chose the diamond shape structure for that use case.

We found no cultural differences regarding the words and VQs selections for the use case of MAR. the top selected word for this use case were *Friendly, reliable*, and *Professional*. But we did find an interesting observation regarding the perception of MAR's role. It seems like there are three different interpretations of medical roles: **Nursing roles** (*Calming, Comforting*, and *Compassionate*), **Expert roles** (*Practical, Clean*, and *Purposeful*), and **Comforting roles** (*Gentle, Familiar*, and *Funny*). Further investigation is needed to understand rather these interpretations are culture related. These would also have implications for the robot's behaviors, as [59] stated:" a robot nurse might enter patients' social spaces, whereas a robot surgeon or a robot janitor in the same hospital won't".

When asked to design an ALR, the designers were concerned with the idea of older adults using technology; they selected the words: *Reliable, Friendly*, and *Inviting*. And avoided the word *Innovative*, which was selected significantly less for this use case. Four designers suggested additional words: *Familiar*, *protecting*, *gentle*, and *easy to use*; all four are related to protecting the elderly. It would be interesting to investigate the perceptions and preferences of residents in assistive living facilities.

To conclude, when designing a new SAR, designers must consider the four layers of the context in the development stage; the selected VQs of a new robot should align with a robotic character suitable to the intended domain, environment, users, and the robot's role. It's recommended to allow a certain level of customization in the implementation stage to provide an easy way to adjust the design to different cultures' perceptions of VQs and robotic roles. This is even more crucial for the case of personal SARs, where personal preferences are more dominant. Figure 16 summarizes these recommendations in a model.

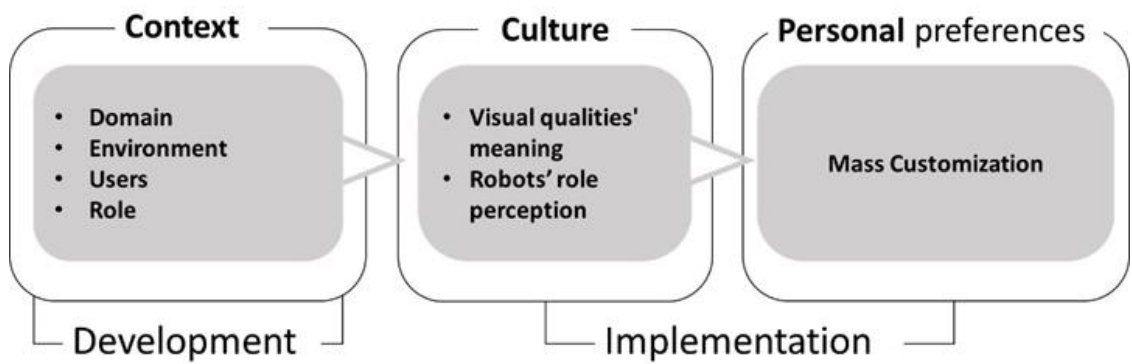

Figure 16: Three factors that must be considered when designing new SARs

This research, however, is subject to several limitations. First, participants were limited with the VQs' selection and could only select three VQs to design their SAR: structure, outline, and color, with a closed set of options. Therefore, all final designs share the same screen, wheels, proportions, and dimensions. Second, the two cultures examined in this study are western and are part of the same aesthetic preference cultural group [49]. Therefore, our subsequent studies will explore more varied cultures and a different set of VQs, such as height preferences that were found to be culturally related [48] and the design of the screen and the wheels that appear to be important to the designers.

We believe our findings will further support the design process of new SARs And form design guidelines for context-based and culturally aware future SARs.

## ACKNOWLEDGMENTS

This research was supported by the Ministry of Innovation, Science and Technology, Israel (grant 3-15625), and by Ben-Gurion University of the Negev through the Helmsley Charitable Trust, the Agricultural, Biological and Cognitive Robotics Initiative, the W. Gunther Plaut Chair in Manufacturing Engineering and by the George Shrut Chair in Human Performance Management.

**APPENDICES**

    A.

| N (%) | Case study | | | | | | Culture | | | Gender | | Design field | | Experience | |
|---|---|---|---|---|---|---|---|---|---|---|---|---|---|---|---|
| | General | COR | ALR | PAR | MAR | | Israeli | German | | Male | Female | 3D | 2D | Students | Prof. |
| Innovative | 72 (36%) | 19 (35.2%) | 8 (16.3%) | 15 (36.6%) | 30 (53.6%) | | 42 (37.5%) | 30 (34.1%) | | 21 (27.3%) | 51 (42.5%) | 35 (33.7%) | 37 (38.5%) | 42 (40.4%) | 30 (31.3%) |
| Inviting | 105 (52.5%) | 21 (38.9%) | 37 (75.5%) | 20 (48.8%) | 27 (48.2%) | | 50 (44.6%) | 55 (62.5%) | | 40 (51.9%) | 63 (52.5%) | 44 (42.3%) | 61 (63.5%) | 56 (53.8%) | 49 (51%) |
| Cute | 33 (16.5%) | 5 (9.3%) | 11 (22.4%) | 8 (19.5%) | 9 (16.1%) | | 21 (18.8%) | 12 (13.6%) | | 14 (18.2%) | 19 (15.8%) | 15 (14.4%) | 18 (18.8%) | 16 (15.4%) | 17 (17.7%) |
| Elegant | 47 (23.5%) | 4 (7.4%) | 11 (22.4%) | 18 (43.9%) | 14 (25%) | | 32 (28.6%) | 15 (17%) | | 18 (23.4%) | 29 (24.2%) | 28 (26.9%) | 19 (19.8%) | 24 (23.1%) | 23 (24%) |
| Massive | 4 (2%) | 3 (5.6%) | 1 (2%) | 0 (0%) | 0 (0%) | | 2 (1.8%) | 2 (2.3%) | | 2 (2.6%) | 2 (1.7%) | 4 (3.8%) | 0 (0%) | 2 (1.9%) | 2 (2.1%) |
| Friendly | 155 (77.5%) | 33 (61.1%) | 43 (87.8%) | 31 (75.6%) | 48 (85.7%) | | 80 (71.4%) | 75 (85.2%) | | 56 (72.7%) | 96 (80%) | 72 (69.2%) | 83 (86.5%) | 78 (75%) | 77 (80.2%) |
| Authoritative | 39 (19.5%) | 26 (48.1%) | 4 (8.2%) | 0 (0%) | 9 (16.1%) | | 34 (30.4%) | 5 (5.7%) | | 13 (16.9%) | 26 (21.7%) | 27 (26%) | 12 (12.5%) | 22 (21.2%) | 17 (17.7%) |
| Aggressive | 1 (0.5%) | 1 (1.9%) | 0 (0%) | 0 (0%) | 0 (0%) | | 0 (0%) | 1 (1.1%) | | 1 (1.3%) | 0 (0%) | 1 (1%) | 0 (0%) | 1 (1%) | 0 (0%) |
| Reliable | 157 (78.5%) | 38 (70.4%) | 43 (87.8%) | 29 (70.7%) | 47 (83.9%) | | 85 (75.9%) | 72 (81.8%) | | 60 (77.9%) | 96 (80%) | 76 (73.1%) | 81 (84.4%) | 87 (83.7%) | 70 (72.9%) |
| Professional | 125 (62.5%) | 37 (68.5%) | 29 (59.2%) | 20 (48.8%) | 39 (69.6%) | | 65 (58%) | 60 (68.2%) | | 41 (53.2%) | 83 (69.2%) | 58 (55.8%) | 67 (69.8%) | 74 (71.2%) | 51 (53.1%) |
| Intelligent | 97 (48.5%) | 26 (48.1%) | 22 (44.9%) | 24 (58.5%) | 25 (44.6%) | | 42 (37.5%) | 55 (62.5%) | | 32 (41.6%) | 64 (53.3%) | 40 (38.5%) | 57 (59.4%) | 57 (54.8%) | 40 (41.7%) |
| Threatening | 4 (2%) | 3 (5.6%) | 0 (0%) | 0 (0%) | 1 (1.8%) | | 3 (2.7%) | 1 (1.1%) | | 1 (1.3%) | 3 (2.5%) | 4 (3.8%) | 0 (0%) | 1 (1%) | 3 (3.1%) |

B

| | | Case study | | | | | Culture | | Gender | | Design field | | Experience | |
|---|---|---|---|---|---|---|---|---|---|---|---|---|---|---|
| N (%) | General | COR | ALR | PAR | MAR | | Israeli | German | Male | Female | 3D | 2D | Students | Prof. |
| A shape | 62 (31%) | 12 (22%) | 18 (37%) | 10 (24%) | 22 (39%) | | 34 (30.5%) | 28 (32%) | 23 (30%) | 39 (32.5%) | 30 (29%) | 32 (33%) | 36 (34.5%) | 26 (27%) |
| Diamond | 61 (31.5%) | 14 (26%) | 16 (33%) | 21 (51%) | 10 (18%) | | 32 (28.5%) | 29 (33%) | 20 (26%) | 39 (32.5%) | 30 (29%) | 31 (32%) | 36 (34.5%) | 25 (26%) |
| Hourglass | 26 (13%) | 6 (11%) | 8 (16%) | 3 (7%) | 9 (16%) | | 13 (11.5%) | 13 (15%) | 11 (14%) | 15 (12.5%) | 12 (11.5%) | 14 (14.5%) | 9 (9%) | 17 (18%) |
| Rectangle | 11 (5.5%) | 3 (5.5%) | 0 (0%) | 3 (7%) | 5 (9%) | | 5 (4.5%) | 6 (7%) | 6 (8%) | 4 (3.5%) | 7 (6.5%) | 4 (4%) | 3 (3%) | 8 (8%) |
| V-shape | 40 (20%) | 19 (35%) | 7 (14%) | 4 (10%) | 10 (18%) | | 28 (25%) | 12 (14%) | 17 (22%) | 23 (19%) | 25 (24%) | 15 (15.5%) | 20 (19%) | 20 (21%) |
| Chamfered | 25 (12.5%) | 11 (20%) | 3 (6%) | 6 (15%) | 5 (9%) | | 16 (14%) | 9 (10%) | 10 (13%) | 14 (12%) | 17 (16%) | 8 (8%) | 16 (15.5%) | 9 (9%) |
| Rounded | 175 (87.5%) | 43 (80%) | 46 (94%) | 35 (85%) | 51 (91%) | | 96 (86%) | 79 (90%) | 67 (87%) | 106 (88%) | 87 (84%) | 88 (92%) | 88 (84.5%) | 87 (91%) |
| Dark colors | 22 (11%) | 5 (9%) | 2 (4%) | 12 (29%) | 3 (5.5%) | | 9 (8%) | 13 (15%) | 10 (13%) | 12 (10%) | 10 (9.5%) | 12 (12.5%) | 11 (11%) | 11 (11.5%) |
| White and blue | 67 (33.5%) | 13 (24%) | 16 (33%) | 16 (39%) | 22 (39%) | | 40 (36%) | 27 (30.5%) | 28 (36.5%) | 39 (32.5%) | 35 (33.5%) | 32 (33.5%) | 34 (32.5%) | 33 (34.5%) |
| White | 111 (55.5%) | 36 (67%) | 31 (63%) | 13 (32%) | 31 (55.5%) | | 63 (56%) | 48 (54%) | 39 (50.5%) | 69 (57.5%) | 59 (57%) | 52 (54%) | 59 (56.5%) | 52 (54%) |

26